# AUTOMATIC ISP IMAGE QUALITY TUNING USING NONLINEAR OPTIMIZATION

*Jun Nishimura, Timo Gerasimow, Rao Sushma, Aleksandar Sutic, Chyuan-Tyng Wu, Gilad Michael*

Intel Corporation

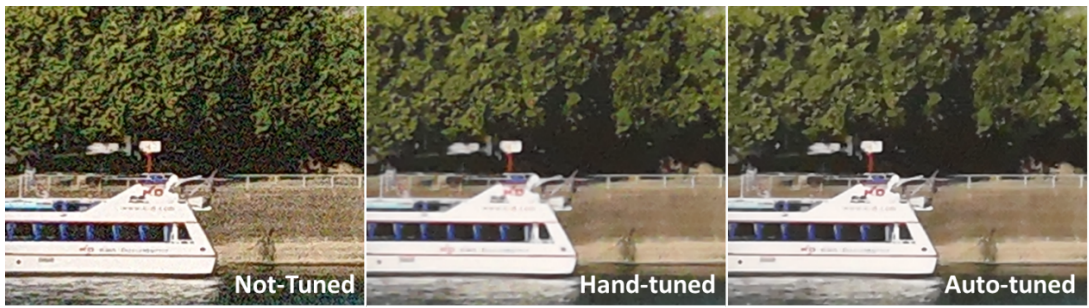

**Fig. 1.** Comparison of ISP output images with different tunings on IMX260 ISO400.
From left to right: Not-tuned, Hand-tuned by image quality expert, Auto-tuned.

## ABSTRACT

Image Signal Processor (ISP) comprises of various blocks to reconstruct image sensor raw data to final image consumed by human visual system or computer vision applications. Each block typically has many tuning parameters due to the complexity of the operation. These need to be hand tuned by Image Quality (IQ) experts, which takes considerable amount of time. In this paper, we present an automatic IQ tuning using nonlinear optimization and automatic reference generation algorithms. The proposed method can produce high quality IQ in minutes as compared with weeks of hand-tuned results by IQ experts. In addition, the proposed method can work with any algorithms without being aware of their specific implementation. It was found successful on multiple different processing blocks such as noise reduction, demosaic, and sharpening.

## 1. INTRODUCTION

The effort required to tune Image Signal Processor (ISP) is tremendously high. In modern ISP, there are tens of processing blocks, and each block can have tens high level tunable parameters. Today, they are tuned manually by Image Quality (IQ) experts for each sensor and use case. Due to the number of parameters to be modified, interdependency between blocks, algorithm complexity, and thorough manual review of image quality on different images, IQ tuning is a highly time consuming process. Therefore, it is a bottleneck in delivering imaging solutions to different customers since each customer will have different use cases with various image sensors.

To our best knowledge, limited prior works exist on automating ISP IQ tuning process. Ernest C. H. Cheung et al. applied Covariance Matrix Adaptation Evolution Strategy (CMA-ES) to optimize Konolige's stereo matching algorithm [1]. Luke Pfister et al. proposed the parameter tuning specific to transform domain noise reduction algorithm [2]. Anish Mittal et al. proposed to train noise sigma parameter of BM3D noise reduction using MS-SSIM measurement with respect to the clean image as a reference [3].

In the industry, AlgoLux provides a solution for automatically optimizing ISP tuning parameters using IQ KPIs (Key Performance Indicators) such as SNR, Texture Acutance, and Sharpness [4]. Because there is no technical paper published on their solution, in order to evaluate similar approach, we have investigated a similar solution using multi-objective optimization evolutionary algorithm (MOEA/D) [5] to stochastically find global optimal ISP tuning for multiple IQ KPIs. There were several difficulties observed in the use of IQ KPIs.

IQ KPIs do not impose enough constraints on image artifacts. Good IQ KPIs can be achieved even in presence of unacceptable artifacts. This is similar to the case when IQ is hand-tuned without subjective evaluation. Such observation has driven us towards our approach of not relying on IQ KPIs.

MOEA/D generates a pareto-front, a set of solutions that takes best trade-offs among multiple objectives. It was difficult to control the smoothness of the pareto-front. Two adjacent tuning sets along the pareto-front [5] can result in very different images, making optimal tuning difficult to achieve. Thirdly, the solution does not take into account the tuning parameter transition smoothness among different sensor gains. In typical imaging system, ISP tuning parameters are stored per certain discrete sensor gains (x1, x2, x4, etc.), and they are interpolated according to the actual sensor gain setting in run-time. If the tuning parameters transition is too abrupt in between the gains, the interpolated tuning parameters may result in an unacceptable image quality.

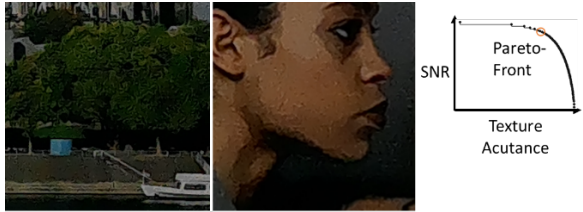

**Fig. 2.** Optimization based on IQ KPIs, resulting in image artifacts. Crops are images by tuning found on pareto-front.

In this paper, we present an automatic ISP IQ tuning using nonlinear optimization and automatic reference generation. By appropriately preparing the reference image for each of key function blocks in ISP, we have found that ISP's high dimensional tuning parameters can be automatically tuned to get high image quality image, which can often outperform IQ experts hand-tuning.

## 2. AUTOMATIC ISP IQ TUNING

The proposed automatic ISP IQ tuning solution is composed of the following components: Nonlinear Optimization, Parameter Abstraction, Automatic Reference Generator, and Fitness Calculation.

Firstly, camera module specific calibration is done. In this step, basic calibrations such as black level, lens shading, linearization and color correction are carried out.

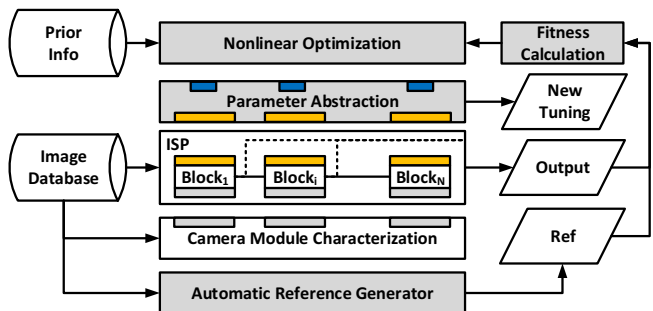

**Fig. 3.** Overview of automatic ISP IQ tuning.

The reference image (Ref) is different for each block. It is orthogonal to the specific ISP algorithm implementation with the exception that it must be better than what is achievable by the specific ISP. Nonlinear optimization algorithm will find the optimal set of tuning parameters which will result in output images with best possible fitness measure against reference. The reference needs to be significantly better than the algorithm being tuned and for that techniques that are not available to the real time ISP.

Nonlinear optimization algorithm does not interact with ISP tuning parameters directly. It is recommended to have a parameter abstraction in case ISP blocks are exposing low level parameters. For example, 5x5 blur kernel coefficients can be efficiently abstracted with a single cut-off frequency parameter. In the parameter abstraction, we have also normalized the range of parameters to [0.0, 1.0] for easy interaction with any nonlinear optimization algorithms. To automatically tune ISP pipeline, each ISP block is optimized with its reference sequentially.

## 3. AUTOMATIC REFERENCE GENEREATOR

In this paper, we have assumed a basic ISP pipeline as shown in Figure 4, and enabled automatic tuning of 4 key blocks: Bayer Noise Reduction, Demosaic, YUV Noise Reduction, and Sharpening. These are key blocks which decide overall ISP IQ and significant amount of tuning effort is put into.

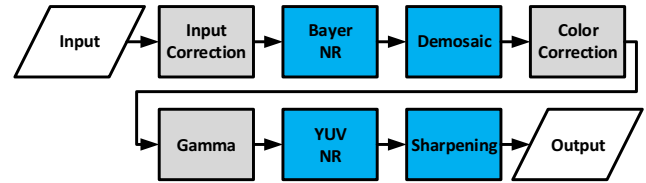

**Fig. 4.** Assumed basic ISP pipeline.

For fitness calculation, simple sum of absolute difference (SAD) between output and reference is used in different domain according to the target block. We have explored other methods such as SSIM and MS-SSIM, but they didn't show significant advantage. We believe this is due to the quality of the reference image.

### 3.1. Reference for Noise Reduction

Achieving good IQ is all about taking a good trade-off between different IQ aspects. For noise reduction, it is detail vs. noise. To break the trade-off, multiple frames of static scene can be temporally averaged to generate good reference for spatial noise reduction. To reflect different customers IQ preference, the number of frames $N$ can be adjusted. For fine control, one can use blending weight of [0.0, 1.0] for 2 frames.

$$I_{fusion} = 1/N \sum_{t=1}^{N} I_i(t)$$

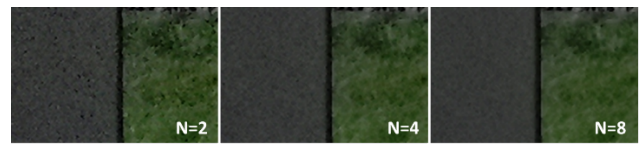

**Fig. 5.** Reference images with different noise reduction strength preference.

### 3.2. Reference for Demosaic

For demosaic, reference RGB image and its corresponding Bayer image can be simulated by adding noise on linearized clean RGB image using pre-calibrated noise model [6], and apply Bayer subsampling as in [7].

### 3.3. Reference for Sharpening

Sharpening is a challenging block which requires a good trade-off between noise and sharpness. With temporal averaging, it is easy to break this trade-off. We used edge directed unsharp masking on temporally averaged frame. First, the gradients $G_{\{H,V\}}$ using Scharr operator is used to calculate weight $w = |G_H|/(|G_H| + |G_V|)$. Directional detail can be calculated as

$$D_{dir} = wD_H \otimes I_{fusion} + (1-w)D_V \otimes I_{fusion},$$

where $D_H = [-1, 2, -1]^T$ and $D_V = D_H^T$. Non-directional detail is calculated as

$$D_{ndir} = I_{fusion} - I_{fusion} \otimes G_{\sigma_{USM}},$$

where Gaussian low pass filter $G_{\sigma_{USM}}$ is configured as 9x9 with $\sigma_{USM} = 2.5$. Two components are blended as

$$I_{ref} = I_{fusion} + \alpha \cdot \alpha_{ndir}(w_{ndir}D_{ndir} + (1 - w_{ndir})D_{dir}),$$

where $w_{ndir} = exp(-min(G_H, G_V)^2/\sigma_{ndir}^2)$ with $\sigma_{ndir} = 0.5$, and $\alpha_{ndir} = 1 - exp(-K \otimes |D_{ndir}|/\sigma_\alpha^2)$. $\sigma_\alpha$ is calculated as mean + standard deviation of $K \otimes |D_{ndir}|$ at flat region. $K$ is 9x9 box filter. $\alpha$ controls an overall sharpness.

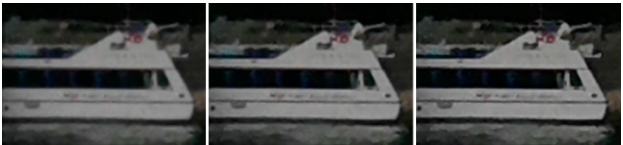

**Fig. 6.** Reference images with different sharpening strength.

## 4. NONLINEAR OPTIMIZATION

Since ISP is not always differentiable by its tuning parameters, gradient-free nonlinear optimization approach is employed. There are many nonlinear algorithms for both global and local optimization. Use of the global optimization such as CMA-ES, PSO, and ABC [8] is good for global exploration in high dimensional search space. However, it has low repeatability in its convergence and it can converge to somewhere close but not onto certain minima. On the other hand, while local optimization, such as Nelder-Mead Simplex [9] and Subplex [9], converge relatively well toward certain minima, its convergence depends on how the optimization is initialized. Therefore, global and local optimization are combined in our approach by initializing local optimization with global optimization results.

First stage of the optimization will carry out the global optimization. This stage will globally explore the high dimensional tuning parameter search space for good candidates. At the second stage, local optimization is carried out to refine the candidates to further look for better fitness measurement.

To interact with tuning parameters, parameter abstraction is added. This step serves 2 purposes. Firstly, it normalizes the range of each tuning parameter to [0.0, 1.0] for easy handling in various nonlinear optimization algorithms. Secondly, some ISP tuning parameters can be too low level with unnecessary flexibility. Since the search space can be highly non-convex, it is recommended to reduce the degree of freedom.

### 4.1. Global Optimization
Artificial Bee Colony (ABC) method is chosen for global optimization [8]. It is swarm-intelligence algorithm with good exploration ability and applicability to numerous practical problems. It is also simple, stable, and not requiring complex parameter tuning.

### 4.2. Local Optimization
Subplex method is chosen for local optimization. It is designed especially for high dimensional search. It decomposes the search space into low dimensional subspaces and applies Nelder-Mead Simplex method [9].

**Subplex Algorithm**

Initialize with **global optimization result** or
Initialize with **tuning result from lower sensor gain**
for regularization
while Termination-criteria not met
    Set step size and subspaces
    for each subspace
        n ← subspace dimension
        do **Nelder-Mead Simplex** method
            Create an initial simplex with n + 1 points
            $Y$ consist of $n + 1$ linearly independent points
            while Termination-criteria not met
                $x_k \leftarrow \arg\max f(x_i), x_i \in Y$
                $\bar{x} \leftarrow \frac{1}{n+1}\sum_{i \in Y \setminus \{x_k\}} x_i$
                $\Delta x \leftarrow x_k - \bar{x}$
                Replace $x_k$ in Y by $x_k^+ \leftarrow \bar{x} + \Delta x$

### 4.2. Prior Information
When optimizing tuning parameters, one can limit search range by appropriately leveraging prior information such as tuning know-hows and algorithm assumptions. In addition, we used tuning setting from lower sensor gains to initialize the local optimization. In this way, the tuning parameter transition between adjacent sensor gains can be regularized.

## 5. EXPERIMENTAL RESULTS

Automatic ISP IQ tuning is evaluated using SONY IMX260 and OV16860 sensors. RAW images were captured at sensor gains from ISO50 to ISO400. For reference image generation, we captured burst of 10 frames as input. For demosaic reference, the dataset proposed by [7] was used for clean images. For global optimization, a population size was set to 40.

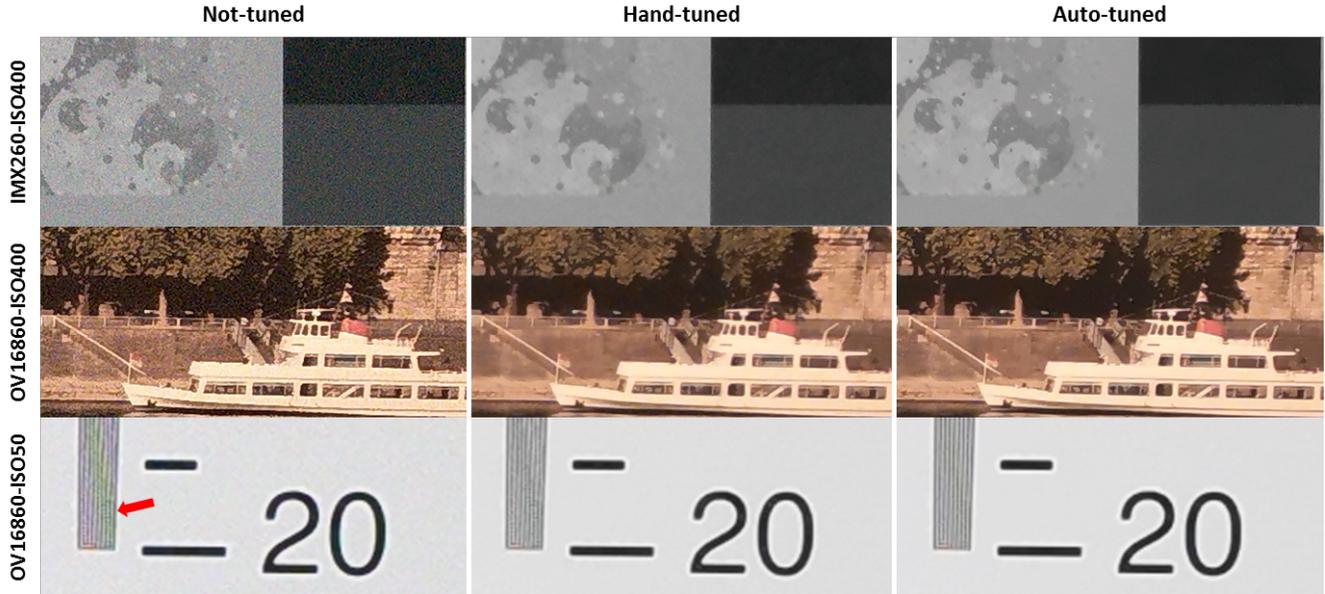

**Fig. 7.** Comparison among 'Not-tuned', 'Hand-tuned' and 'Auto-tuned' images on IMX260 and OV16860.

Figure 7 shows the image comparisons among 'Not-tuned', 'Hand-tuned', and 'Auto-tuned'. With our solution, even small zipper artifact and false color at high frequency (as indicated in red arrow on OV16860-ISO50) is properly addressed. Table 1 shows Mean Absolute Difference (MAD) in 8bit scale, SSIM, and MS-SSIM [10] between reference and different tunings of each key ISP block. 'Auto-tuned' provides significant IQ improvement over 'Not-tuned' similar to 'Hand-tuned' but with significantly limited tuning time. Typically, automatic tuning process requires 10~20mins per block depending on the size of images used to optimize.

**Table 1**. Difference between reference and each tuning output for each key ISP block.

|  | MAD | | | SSIM | | | MS-SSIM | | |
|---|---|---|---|---|---|---|---|---|---|
|  | Not | Hand | Auto | Not | Hand | Auto | Not | Hand | Auto |
| Bayer NR | 1.7507 | 1.1998 | **1.1618** | 0.8607 | 0.9313 | **0.9357** | 0.9062 | 0.9401 | **0.9421** |
| Demosaic | 11.7463 | 11.3349 | **10.4392** | 0.8968 | 0.9100 | **0.9237** | 0.9728 | 0.9775 | **0.9794** |
| Sharpening | 3.1719 | 1.6488 | **1.4921** | 0.9557 | 0.9822 | **0.9848** | 0.9834 | 0.9958 | **0.9969** |
| YUV-NR | 0.8788 | 0.8781 | **0.8554** | 0.9934 | 0.9952 | **0.9970** | 0.9945 | 0.9961 | **0.9971** |

### 5.1. Repeatability

Global optimization can often result in different convergence among multiple runs due to its stochastic nature. Table 2 compares MAD in 8bit scale for 10 runs of Bayer NR automatic tuning with different optimization flows. By combining local optimization with global optimization, fitness measure and its variance are reduced. By leveraging the prior information such as certain reasonable tuning parameter ranges to reduce the search space, fitness measure and its variance can further be reduced.

### 5.2. Parameter transition regularization

Ensuring smooth tuning parameter transition between adjacent sensor gains is important for dynamic imaging system. Figure 8 shows parameter transition of Bayer noise reduction block between different sensor gains. Notice that by regularizing the optimization, parameters among ISO50 to ISO200 are smoothly transitioning (blue bars). Without the regularization, tuning parameters vary more abruptly (orange bars).

**Table 2.** Average and standard deviation of fitness measure.

|  | Global | Global → Local | Global → Local w/ Prior |
|---|---|---|---|
| AVE | 1.1926 | 1.1793 | **1.1629** |
| STD | 8.34E-03 | 4.48E-03 | **1.70E-03** |

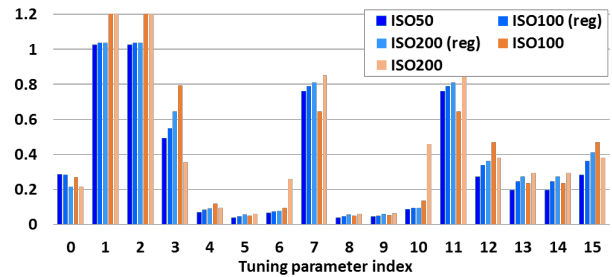

**Fig. 8.** Tuning parameter transition among different gains.

### 6. CONCLUSION

We demonstrated automatic ISP IQ tuning using nonlinear optimization and automatic reference image generation. Using our solution, IQ tuning can be done in minutes and often outperform weeks of manual tuning, and auto-tune often outperforms hand-tuned results. IQ experts use the automatic tuning by specifying IQ preferences and prior information to further enhance the results. Our solution is agnostic to the specific algorithm implementations, and it can be applied to different generation of ISPs.